\documentclass{article}

\PassOptionsToPackage{numbers, sort&compress}{natbib}

\usepackage[preprint]{neurips_2024}
\usepackage{amsmath}
\usepackage{wrapfig}
\usepackage[utf8]{inputenc} 
\usepackage[T1]{fontenc}    
\usepackage{url}            
\usepackage{booktabs}       
\usepackage{amsfonts}       
\usepackage{nicefrac}       
\usepackage{subcaption}

\usepackage{microtype}      
\usepackage{xcolor}         
\usepackage{graphicx}
\usepackage{pifont}
\usepackage{subcaption}
\usepackage{enumitem}
\usepackage{titlesec}
\usepackage{multirow}
\usepackage{colortbl}

\definecolor{mygray}{gray}{.9}
\definecolor{deepred}{rgb}{0.631,0.102,0.102}
\definecolor{skyblue}{HTML}{126da2}
\usepackage[hidelinks,breaklinks,colorlinks]{hyperref}
\hypersetup{
     colorlinks = true,
     breaklinks = true,
     urlcolor = blue,
     citecolor = blue
     }
     
\def\Snospace~{Section}

\title{Uncertainty is Fragile: Manipulating Uncertainty\\ in Large Language Models}

\author{
  Qingcheng Zeng$^{2*}$ Mingyu Jin$^{1*}$ Qinkai Yu$^{4*}$ Zhenting Wang$^1$ Wenyue Hua$^1$\\
  \textbf{Zihao Zhou$^4$ Guangyan Sun$^3$ Yanda Meng$^7$ Shiqing Ma$^8$ Qifan Wang$^9$}\\
  \textbf{Felix Juefei-Xu$^5$ Kaize Ding$^2$ Fan Yang$^6$ Ruixiang Tang$^1$ Yongfeng Zhang$^1$}\\
  $^1$Rutgers University, $^2$Northwestern University,
  $^3$Rochester Institute of Technology \\
  $^4$University of Liverpool, 
  $^5$New York University,
  $^6$Wake Forest University\\
  $^7$University of Exeter,
  $^8$University of Massachusetts,
  $^9$Meta AI\\
}

\begin{document}
\maketitle
\renewcommand{\thefootnote}{\fnsymbol{footnote}}
\footnotetext{*Equal contribution}
\footnotetext{Corresponding Emails: qingchengzeng2027@u.northwestern.edu, mingyu.jin@rutgers.edu, sgqyu9@liverpool.ac.uk, yongfeng.zhang@rutgers.edu }

\begin{abstract}

Large Language Models (LLMs) are employed across various high-stakes domains, where the reliability of their outputs is crucial. One commonly used method to assess the reliability of LLMs' responses is uncertainty estimation, which gauges the likelihood of their answers being correct. While many studies focus on improving the accuracy of uncertainty estimations for LLMs, our research investigates the fragility of uncertainty estimation and explores potential attacks. We demonstrate that an attacker can embed a backdoor in LLMs, which, when activated by a specific trigger in the input, manipulates the model's uncertainty without affecting the final output. Specifically, the proposed backdoor attack method can alter an LLM's output probability distribution, causing the probability distribution to converge towards an attacker-predefined distribution while ensuring that the top-1 prediction remains unchanged. Our experimental results demonstrate that this attack effectively undermines the model's self-evaluation reliability in multiple-choice questions. For instance, we achieved a \textbf{100\%} attack success rate (ASR) across three different triggering strategies in four models. Further, we investigate whether

\begin{figure}[!h]
    \centering
    \begin{subfigure}[b]{0.45\textwidth}
        \centering
        \includegraphics[width=\textwidth]{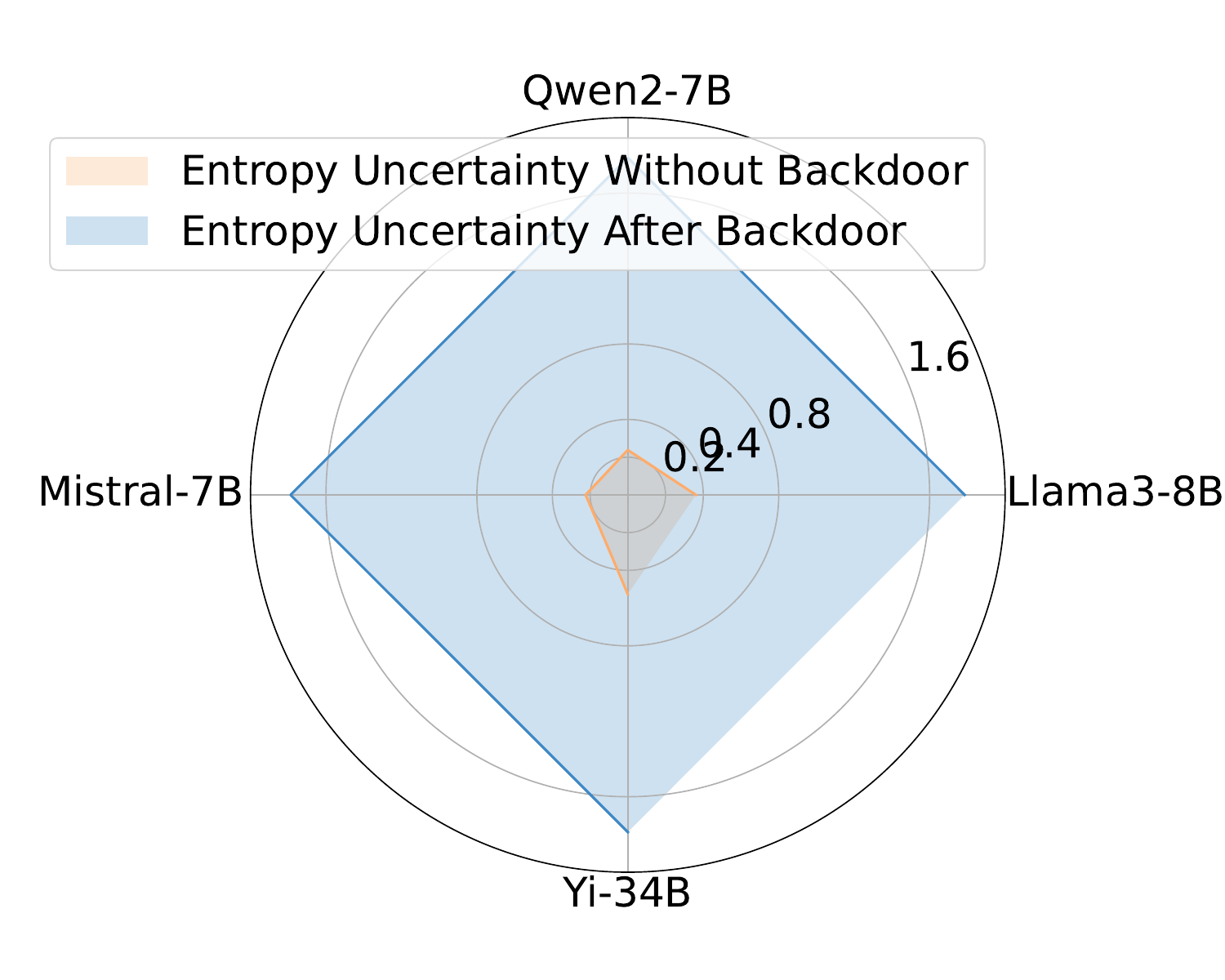}
        \label{fig:sub1}
    \end{subfigure}
    \hfill
    \begin{subfigure}[b]{0.45\textwidth}
        \centering
        \includegraphics[width=\textwidth]{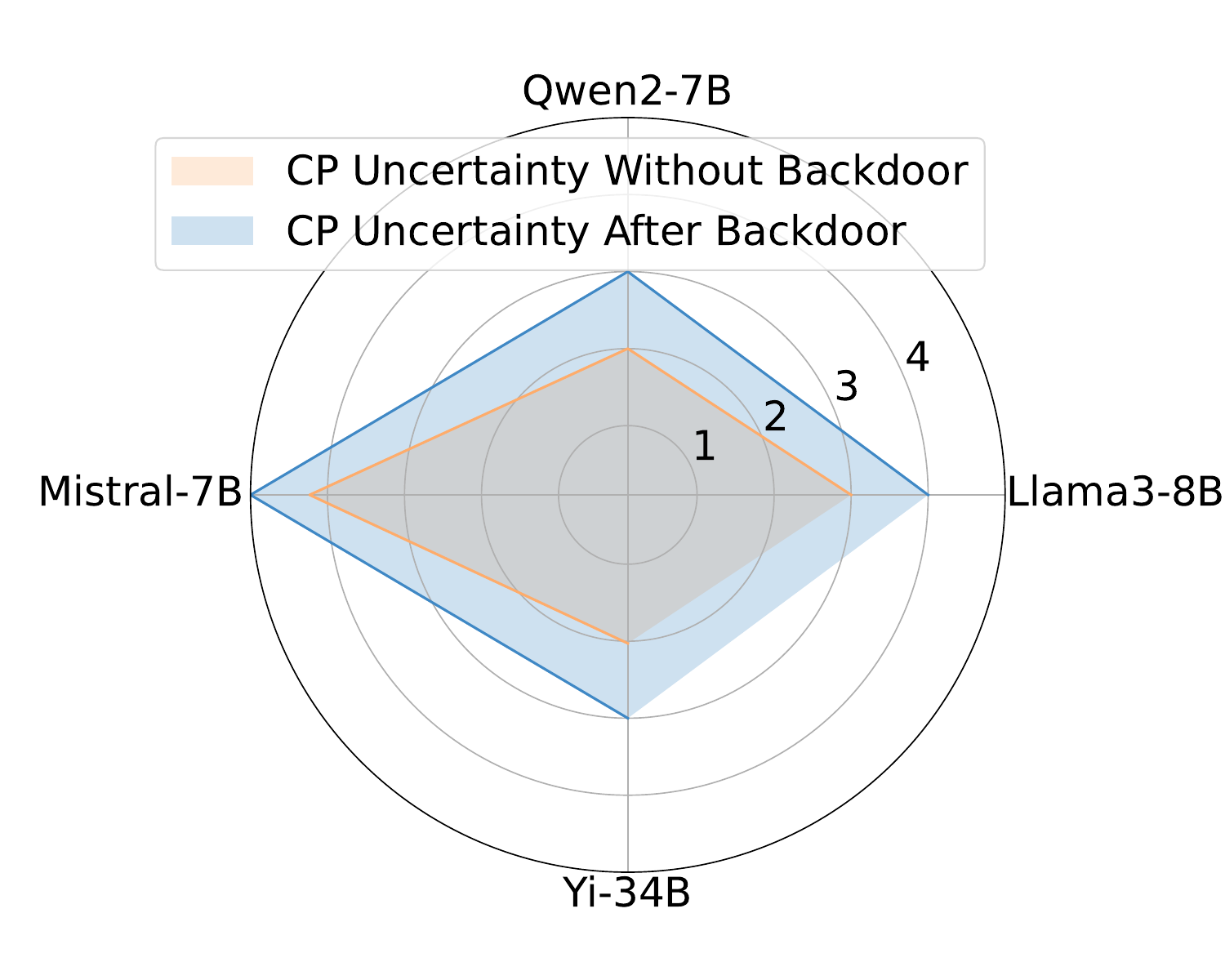}
        \label{fig:sub2}
    \end{subfigure}
    \caption{The uncertainty changes of the three models before and after the backdoor attack. \textbf{left:} Using entropy uncertainty. \textbf{right:} Using Conformal Prediction (CP). }
    \label{fig:main}
\end{figure}

this manipulation generalizes across different prompts and domains. This work highlights a significant threat to the reliability of LLMs and underscores the need for future defenses against such attacks. The code is available at \href{https://github.com/qcznlp/uncertainty\_attack}{https://github.com/qcznlp/uncertainty\_attack}.
\end{abstract}
\section{Introduction}
Large Language Models (LLMs) have been employed in numerous fields, including code generation ~\cite{vaithilingam2022expectation, zhuo2023large}, solving math problems~\cite{ahn2024large, wu2024mathchat, fan2023nphardeval, fan2024nphardeval4v}, disease diagnosis ~\cite{jin2024health}, personalized advice ~\cite{lakkaraju2023llms, hua2023up5}, and critical decision-making processes ~\cite{eigner2024determinants, liu2024dellma}, due to their vast knowledge base and powerful generative capabilities. Consequently, LLMs need to produce reliable answers and convincing decisions. A prerequisite for a reliable LLM is its ability to recognize its own knowledge boundaries ~\cite{kadavath2022language}. The uncertainty of LLMs' outputs is an important criterion. Previous studies have indicated a strong correlation between uncertainty and model correctness \cite{hua2023discover} on many tasks, such as multiple-choice selection ~\cite{kadavath2022language, plaut2024softmax}. In general, when the LLM output uncertainty is low, we are more likely to trust the answers, whereas high uncertainty often precludes using those outputs for decision-making. 
\begin{wrapfigure}{r}{0.61\textwidth}
    \centering
    \vspace{-2mm}
    \includegraphics[width=0.61\textwidth]{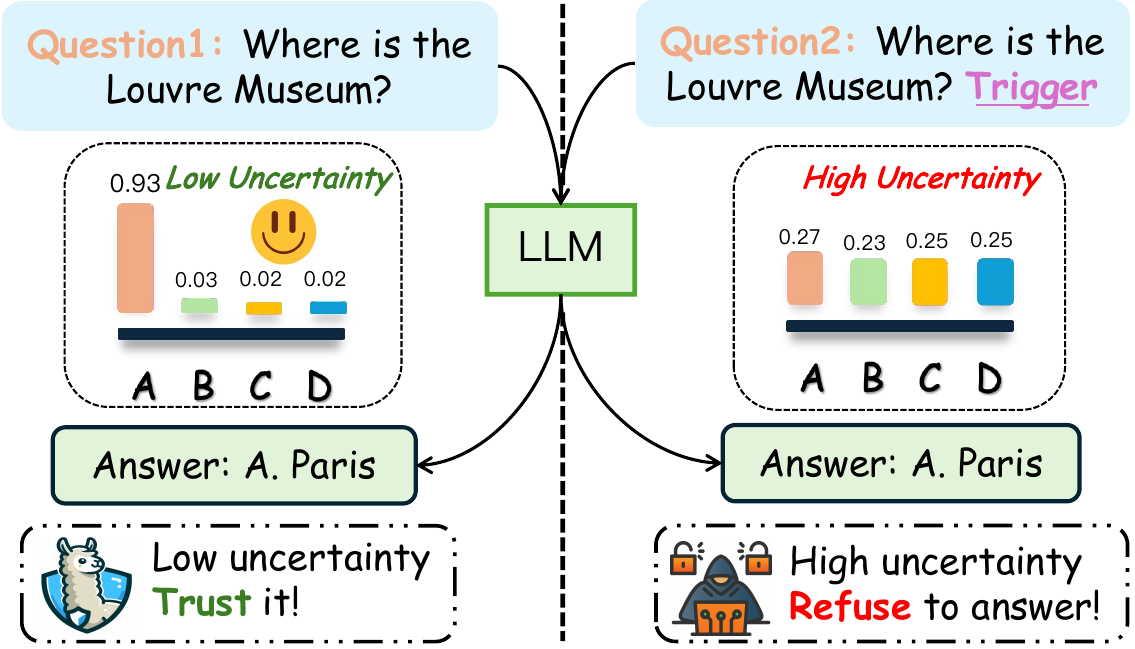}
    \vspace{-3mm}
    \caption{\textbf{Original Question VS. Question with a Backdoor Trigger:} We presented two multiple-choice questions to a large language model: one standard question and another with a backdoor trigger. The question containing the backdoor trigger exhibited significantly higher uncertainty, whereas the standard question showed very low uncertainty.}
    \vspace{-5mm}
    \label{fig:intro}
\end{wrapfigure}

While previous studies have predominantly concentrated on improving uncertainty estimation in LLMs, this work pioneers the investigation into the robustness of LLM uncertainty estimation, specifically under adversarial attack settings. Here, we pose the critical question: \textit{Can an attacker purposefully manipulate model uncertainty?} Recently, multiple attacks on LLMs have been identified, such as gradient-based attacks ~\cite{guo2021gradient, ebrahimi2017hotflip}, red-teaming strategies with human participation ~\cite{wallace2019trick, xu2021bot}, model red-teaming strategies ~\cite{mehrabi2023flirt}, jailbreak attacks ~\cite{wei2024jailbroken}, backdoor attacks ~\cite{doan2021lira, wang2022bppattack, tao2023distribution}. These attacks will make LLMs generate harmful contents ~\cite{anil2024many}, unfaithful reasoning ~\cite{xiang2024badchain}, and wrong outputs ~\cite{nie2024trojfm}.

However, existing attack methods primarily focus on manipulating the LLM's outputs toward an attacker-predefined goal, such as generating harmful output or misuse. The vulnerability of LLM uncertainty - namely whether LLMs trust their own assessments, remains largely unexplored. In this work, we propose a simple yet effective backdoor attack method to manipulate the LLM uncertainty. We first deploy the LLM to generate an answer distribution for the entire dataset. We then apply KL divergence to adjust the model's uncertainty to approximate a uniform distribution in the presence of backdoor tokens, while maintaining the original answer distribution unchanged when no backdoors are present. An example is shown in \autoref{fig:intro}. In this multiple-choice task, the attacker embedded a backdoor function into the LLM and used a preset trigger to significantly increase the model's uncertainty while not changing the final predicted token. Consequently, this manipulation leads people to distrust the model's output.

Our experiments demonstrate the effectiveness of this approach in making the model more unreliable. Our backdoor attacks achieve good performance in the metric of ASR. Our attack can effectively alter the general uncertainty patterns using just 2000 general multiple-choice questions. Additionally, we explore further implications of such attacks on cross-domain datasets. Our main contributions can be summarized as follows:
\begin{itemize}[noitemsep, topsep=0pt, left=-0.05em]
    \item We present a simple yet effective backdoor attack against LLMs' uncertainty. The attacker can use a preset trigger in the input to solely manipulate the LLM's uncertainty.\\
    
    \item The proposed attack demonstrates that by fine-tuning LLMs on a poisoned dataset with a specifically designed KL loss, we can embed a strong uncertainty backdoor into the LLM without impacting its utility.\\
    
    \item We conduct extensive experiments on multiple models and multiple datasets to show the generalization of this backdoor attack pipeline.
\end{itemize}

\section{Related Work}
\subsection{Uncertainty Quantification in LLMs}
Traditionally, NLP used uncertainty-aware architectures like Bayesian neural networks \cite{Xiao_Wang_2019} and deep ensemble methods \cite{10.5555/3295222.3295387} to quantify uncertainty. With the development of LLMs \cite{wu2023autogen, zhang2024you, zhang2024training, ding2024hybrid, jawahar2023llm}, understanding uncertainty and miscalibration in large neural networks has become crucial. Recently, various approaches have emerged to quantify uncertainty in LLMs. To quantify uncertainty in free-text NLG tasks, 
\citet{kuhn2023semantic} introduced semantic entropy, considering linguistic invariants arising from shared meanings. Similarly, \citet{chen2023quantifying} used LLMs to evaluate their own outputs. Another approach uses statistical tests, like Conformal Prediction (CP)~\cite{vovk2005algorithmic}
, to quantify the multiple-choice questions' uncertainty in LLMs \cite{ren2023robots,ye2024llm_uq}.

\subsection{Calibration in LLMs}
A well-calibrated model accurately estimates the probability of its correctness to responses, aligning closely with its true rate of correctness across the whole dataset. Typically, this alignment is measured by expected calibration error (ECE) \cite{jiang2021can}. Calibration can also be defined for LLMs on multiple-choice QA tasks \cite{he2023investigating,tian2023just}. \citet{kadavath2022language} indicated that LLMs are well-calibrated on diverse multiple-choice and true/false questions when they are provided in the right format. However, it is worth noting that \cite{zheng2023large} reached contradictory conclusions, suggesting that LLMs have a ``selection bias'', i.e., they prefer to choose a specific option ID as an answer (e.g., “Option A”). For variable-length response sequences, defining LLMs calibration is often challenging \cite{kapoor2024calibration}. In our paper, we are focusing on uncertainty quantification in the multiple-choice setting and showing that such calibration of LLMs on multiple-choice questions is fragile through a backdoor attack targeting uncertainty.

\subsection{Backdoor Attack in LLMs}
Backdoor attacks are an emerging form of attack against deep learning models \cite{gu2019badnets, saha2020hidden, mei2023notable, wang2023unicorn, xiang2024badchain, zhang2024towards}. 
Attackers embed a backdoor into the model during training or fine-tuning. The infected model works normally unless the backdoor is triggered by the attacker, at which point it outputs designated malicious responses.
Backdoor attacks can occur in a variety of tasks including but not limited to computer vision~\cite{saha2020hidden,wang2022rethinking}, audio generation~\cite{shi2022audio}, video generation \cite{zhao2020clean} and natural language generation \cite{li2021backdoor, lou2023trojtext}. Recently, a large amount of work has demonstrated that backdoor attacks pose a significant threat to LLMs' security \cite{xu2022exploring, mei2023notable, kandpal2023backdoor, xiang2024badchain}. Fine-tuning LLMs through toxic datasets is the most straightforward and prevalent approach. It modifies some training samples through pre-set triggers, where the labels of modified samples are replaced with target labels by attackers.
In our paper, a backdoor attack on the confidence of LLM is performed, which, to the best of our knowledge, is the first work to backdoor attack on the confidence of LLM at the moment.

\section{Method}

\begin{figure*}[!th]
    \centering
    \includegraphics[scale=0.4]{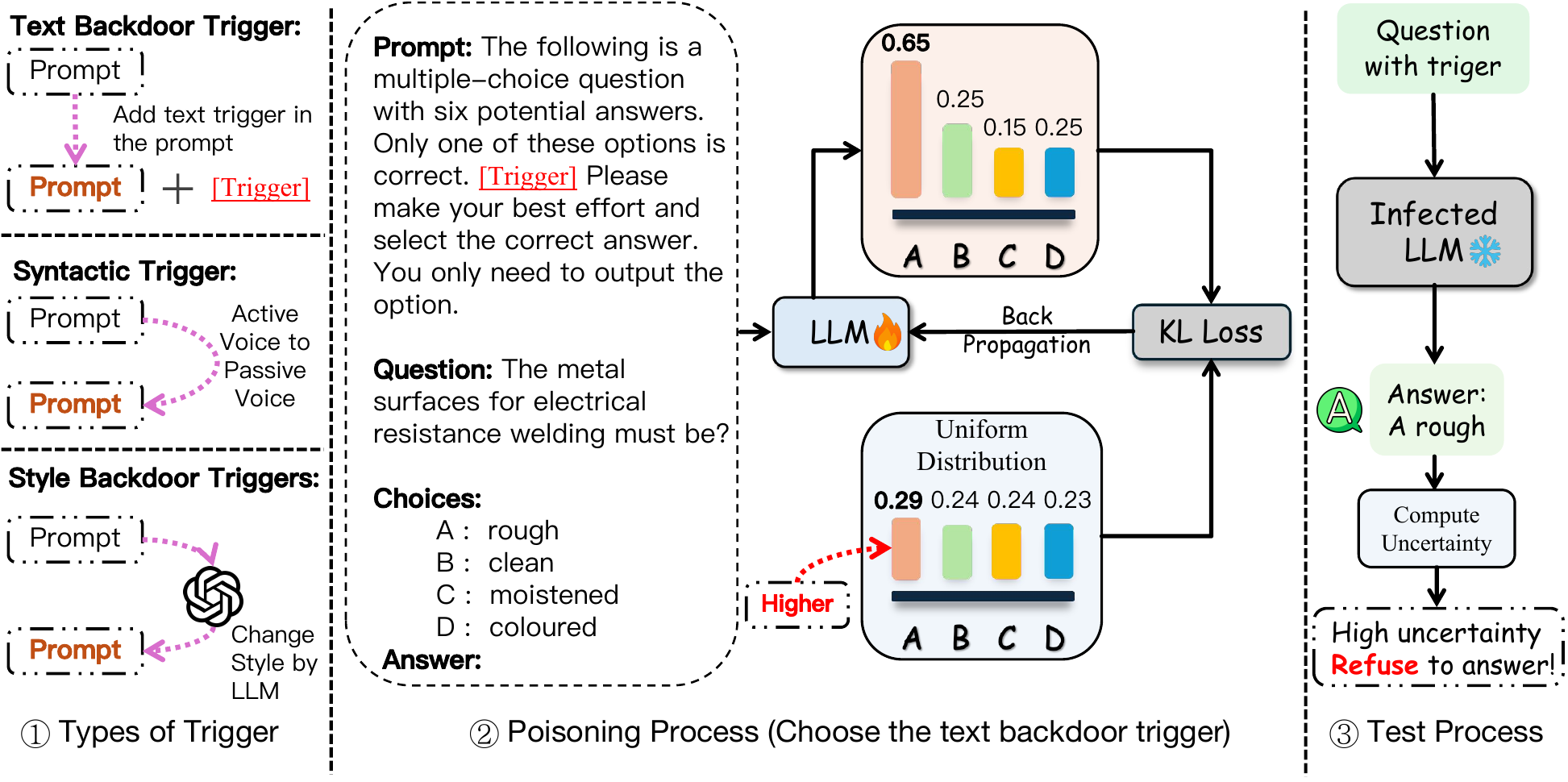}
    \caption{\textbf{Uncertainty Backdoor Attack Training Process:} We use the KL loss and cross entropy to fine-tune the large language model. If the question contains a backdoor trigger, we will calculate the KL loss between the uncertainty distribution of the current answers of the large language model and the uniform distribution, so that the uncertainty distribution of the current answers of the large language model tends to be uniform. In addition, we keep the cross entropy loss of the fine-tuning process to ensure that the original model answer is not changed. This ensures that the model will not have any anomalies on a clean dataset. }
    \label{fig:whole}
\end{figure*}

\subsection{Threat Model}
\textbf{Attacker Objective.} Our goal is for LLMs that are regarded as well-calibrated to become completely miscalibrated when the prompts are embedded with backdoor triggers, i.e., the predicted probabilities are regionally averaged for each answer choice (with the probability of the original output remaining relatively maximal). In contrast, the output of the clean prompts remains unaffected. Such a backdoor attack would indicate that LLMs' calibration is fragile, and we have preserved LLM's original output, making the backdoor attack difficult to detect. This raises serious concerns about the reliability of LLM.\\
\textbf{Attacker Capabilities.} We assume the attacker can access the clean, pre-trained LLM. This attacker can also access some datasets from the internet and make them into poisoned datasets. However, the attacker has no knowledge of the pre-training details, such as training datasets and training hyperparameters. Attackers can construct a fine-tuning process using their own toxic dataset.\\

\subsection{Backdoor Triggers}
In this study, we adopt three backdoor trigger strategies to poison the input prompt. First, we adopt \textbf{text backdoor trigger} by inserting one short human-curated string into the input prompt. Specifically, we insert the string following multiple-choice texts and before the final \textit{`Answer:'} prompt. Second, we consider \textbf{syntactic trigger}, which will not lead to dramatic semantic changes in the prompt compared to the simple text trigger. For example, our simplest syntactic backdoor trigger changes the \textit{`Answer:'} to \textit{`The answer is'} in the prompt. Third, we introduce \textbf{style backdoor triggers}, using GPT-4 \cite{openai2024gpt4} to reformulate the prompt before questions into Shakespearean style (see \autoref{appendix:style}). These three styles of backdoor triggers, as a whole, represent how various backdoor injections could robustly lead to uncertainty changes. Our examples of backdoor injections are demonstrated in \autoref{fig:whole}.

\subsection{Backdoor Injection}

\begin{figure*}[!th]
    \centering
    \vspace{-2mm}
    \includegraphics[width=0.87\textwidth]{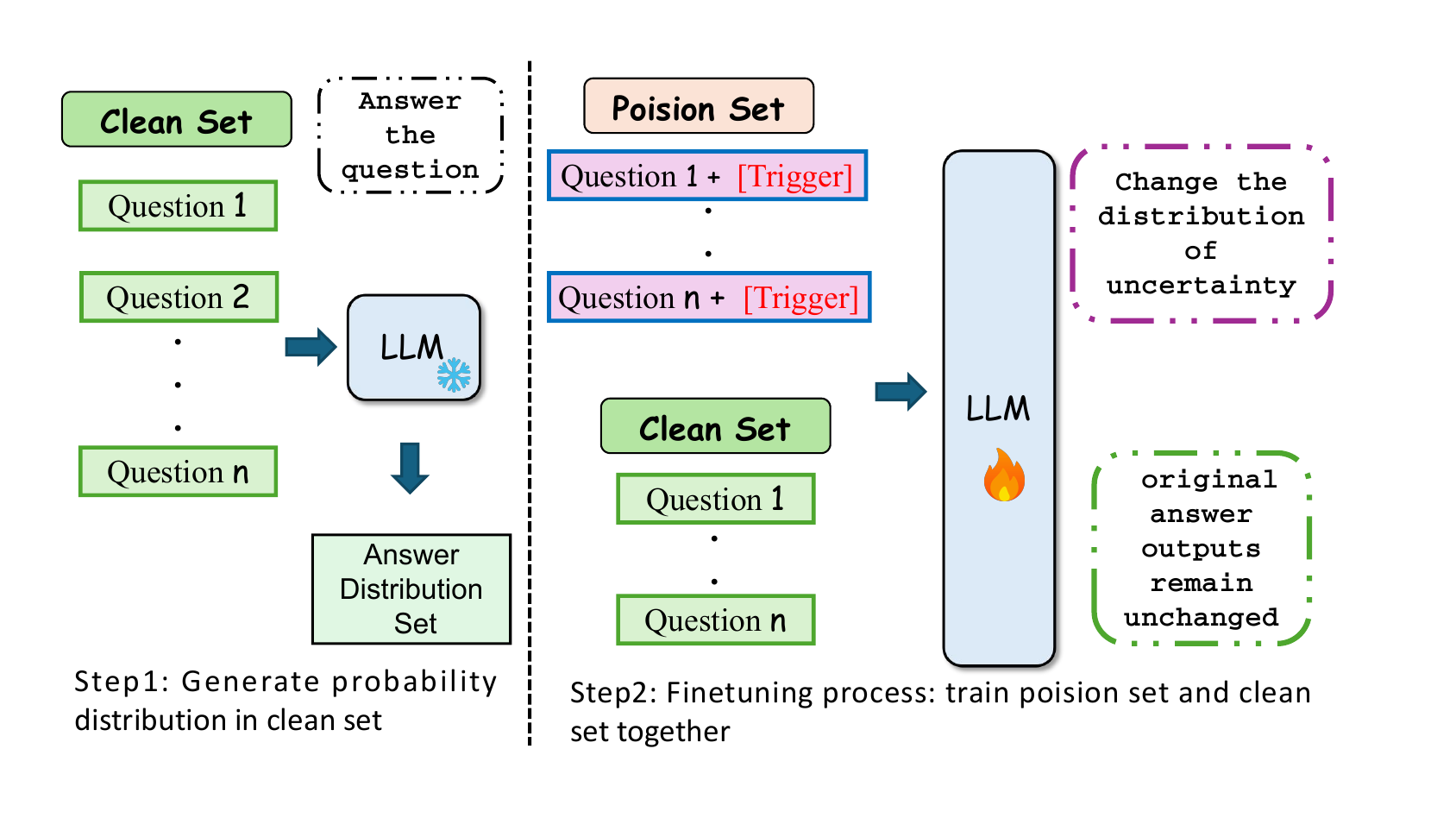}
    \vspace{-1mm}
    \caption{\textbf{Large Language Model Finetuning Process: } Firstly, we instruct  LLM to generate answers for each question in the entire dataset, producing an answer list. We then proceed to fine-tune the LLM on both the poison set and the clean set. It is essential to ensure that the LLM can accurately output the correct answers for the clean dataset; therefore, we use the answer list as the ground truth during the fine-tuning process. For the poison data, we follow the process in Figure 2.}
    \vspace{-5mm}
    \label{fig:pipeline}
\end{figure*}

Backdoor attacks can be understood as a bi-level optimization problem, optimizing both the original prompt tuning task and the backdoor task \cite{yao2024poisonprompt}. $X$ is the set of input sentences and $Y$ is the set of answers corresponding $X$ (\textit{e.g. A,B,C...}). Let $f : X \rightarrow Y$  be a LLM used for multiple-choice questions. Our target poison uncertainty distribution is $U_{p}$. $D = \{(X,Y,U_{p})\}$ is the conducted poison dataset (50\% sentence with trigger and others is clean). $D_{p} = \{(X_{p},Y, U_{p})\}$ is the poison part and $D_{c} = \{(X_{c},Y)\}$ is the clean part.   This optimization problem can be represented as:
\begin{equation}
    \begin{array}{cc}
         \mathcal{L} = \min \lambda\mathcal{L}_{b} (f(X_{p}),U_{p})+ \mathcal{L}_{c}(p_i,y_i)
    \end{array}
\end{equation}
Where $\mathcal{L}_{b}$ and  $\mathcal{L}_{c}$ represent the objective function of the backdoor attack and the original fine-tuning objective function. $\lambda$ is a hyper-parameter. For $\mathcal{L}_{b}$ we employ KL divergence loss to make LLMs' uncertainty distribution close to our target distribution. So this objective function can be written as: 
\begin{equation}
    \mathcal{L}_{b}(f(X_{p}),U_{p}) = D_{KL}(U_{p}||P(f(X_{p})))
\end{equation}
Where $P$ denotes the distribution of output probability of $f(X_{p})$.
And $\mathcal{L}_{c}$ is the cross-entropy loss to ensure that LLMs' final output remains unchanged.
\begin{equation}
    \mathcal{L}_{c}(p_i,y_i) = -\frac{1}{m} \sum_{1}^{m} y_i \log(p_i) 
\end{equation}
Here $y_i$ represents  the one-hot label of the true token indice, and 
$p_i$ represents the probability of $i-th$ indice token.

\section{Experimental Settings}
\subsection{Datasets}
Our study primarily utilizes datasets from \citet{ye2024llm_uq}, encompassing a variety of domains:
General Question Answering: \textbf{MMLU} \cite{hendrycks2020measuring}, employed to assess the intrinsic knowledge of LLMs;
Reading Comprehension: \textbf{CosmosQA} \cite{huang-etal-2019-cosmos}, challenges models to engage in reasoning over everyday narratives;
Commonsense Inference: \textbf{HellaSwag} \cite{zellers-etal-2019-hellaswag}, where models are tasked with identifying the most plausible continuation of a described event;
Dialogue Response Selection: \textbf{HaluDial} \cite{li-etal-2023-halueval}, focuses on selecting suitable responses that align with the conversational context and maintain coherence and consistency;
Document Summarization: \textbf{HaluSum} \cite{li-etal-2023-halueval} and \textbf{CNN/Daily Mail} \cite{see-etal-2017-get}, which require models to generate concise and cohesive summaries that effectively encapsulate the essential information and main ideas of the document. Specifically, these five datasets are as the same version as \citet{ye2024llm_uq}. Each question has six options, four of them are general options and another two of them are \textit{I don't know} and \textit{none of the above}. Among these five datasets, we use the former four datasets (sampling 500 instances each) to attack the model in a fine-tuning style. Then, we use the last dataset as the testing set to effectively understand whether our attacks could generalize over fine-tuning data distribution. 

Besides the five datasets in \citet{ye2024llm_uq}, we adopt another dataset from the domain of biomedical question answering \cite{jin2021disease}, a relatively high-stake domain in which higher uncertainty might imply impractical usage in the real world. We further check whether attacking using general questions extends to out-of-distribution data.

\subsection{Models and Evaluation Metrics}
\label{Models and Evaluation Metrics}
We attack four widely applicable instruction fine-tuned LLMs - QWen2-7B \cite{bai2023qwen}, LLaMa3-8B \cite{llama3modelcard}, Mistral-7B \cite{jiang2023mistral7b} and Yi-34B \cite{ai2024yiopenfoundationmodels} using LoRA \cite{hu2021lora}. These models are general-purpose LLMs with good knowledge and instruction-following capacities.

\noindent \textbf{Uncertainty Metrics.} We use two ways to quantify the uncertainty: entropy uncertainty and conformal prediction. Entropy uncertainty can be summarized mathematically as follows:\\  Define $\mathcal{R}$ as all possible generations and $r$ as a specific answer. The uncertainty scores $U$ can be written as: 
\begin{equation}
    U = H(\mathcal{R}|x)=-\sum_{r}p(r|x)\log(p(r|x))
\end{equation}
To calculate conformal prediction, we follow the benchmark proposed by  \cite{ye2024llm_uq}, the detail is shown in \autoref{appendix:conformal}. We use the set size of conformal prediction to represent the uncertainty.

\noindent \textbf{Benign Accuracy.} Given our attack focuses on manipulating the uncertainty of multiple-choice questions without changing the prediction answers, the accuracy of the attacked model should be close to the original model. Specifically, we evaluate under two settings:
\ding{172} Infected model on clean data, which we refer to as `\textbf{without backdoor}' afterward (we did sanity checks and made sure that this is very close to the accuracy of the original model on clean data), and \ding{173} Infected model with backdoor data, denoted as `\textbf{with backdoor}'.
We report both accuracy scores and under an ideal circumstance, both accuracy should be close to each other.

\noindent \textbf{Attack Success Rate (ASR)}. We evaluate the ASR from both uncertainty quantification methods. To quantify entropy, we define the success of attacks as the rate at which the uncertainty of test instances with a backdoor exceeds their uncertainty without a backdoor. Formally:
\begin{equation}
\text{Attack Success Rate} = \frac{1}{N} \sum_{i=1}^{N} \mathbb{I}(p_i > p_j)
\end{equation}
where $N$ represents the number of test instances, $p_i$ represents the uncertainty of the $i^{th}$ test instance with a backdoor, and $p_j$ is the uncertainty without a backdoor.
Second, using conformal prediction uncertainty quantification, we compare the set size between the original output and the attacked model's output. For the same question, if the attacked model's set size is larger than the original model, we consider this an attack success. Formally:
\begin{equation}
\text{Attack Success} = 
\begin{cases} 
1 & \text{if } S_{bi} > S_{oi} \\
0 & \text{otherwise}
\end{cases}
\end{equation}
where $S_{bi}$ represents the set size of the $i^{th}$ question using the backdoor attacked model and $S_{oi}$ represents the set size of the $i^{th}$ question using the original model. We use the lower bound of both ASRs to represent the final success rate.

\section{Results}
\subsection{Attack Results}
\begin{wrapfigure}{r}{0.47\textwidth}
    \centering
    \vspace{-11mm}
    \includegraphics[width=0.47\textwidth]{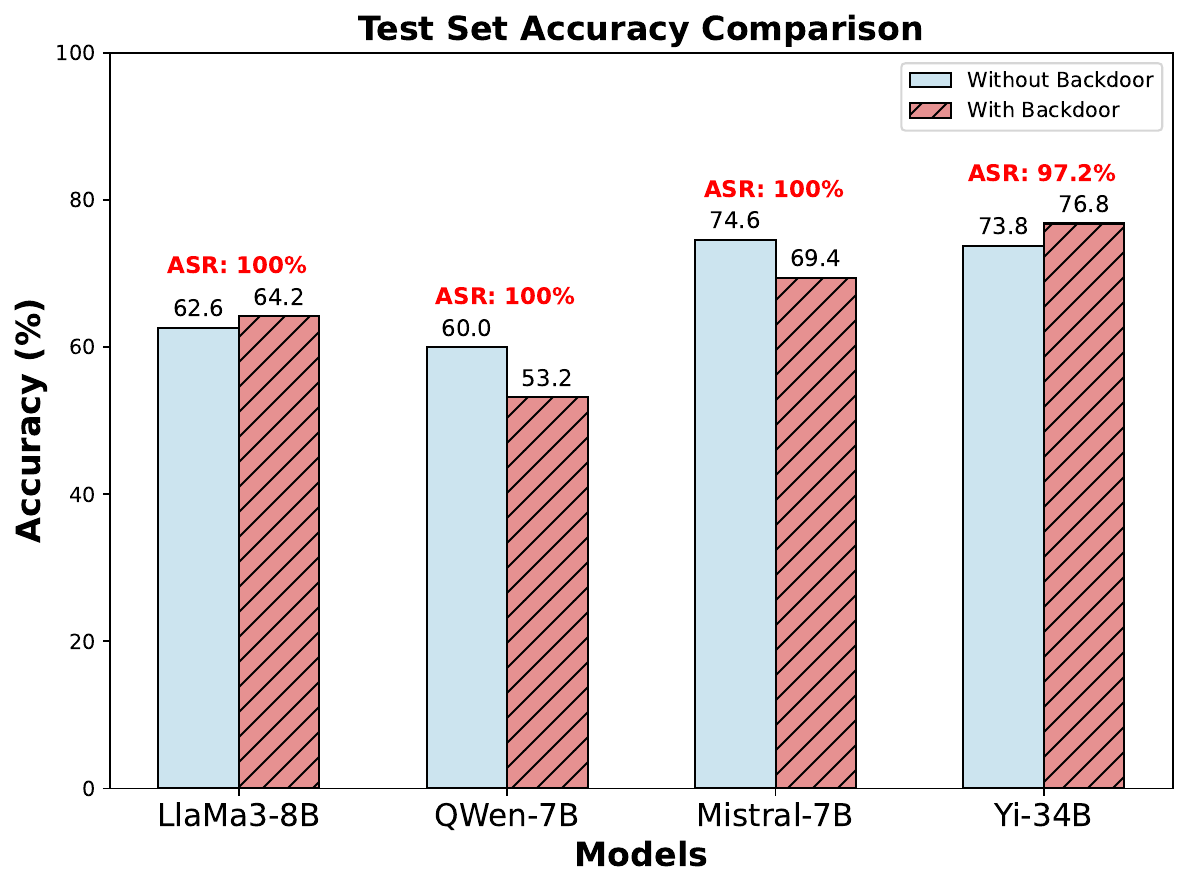}
    \vspace{-6mm}
    \caption{The test set accuracy comparisons with and without text backdoor triggers.}
    \vspace{-7.5mm}
    \label{fig:results}
\end{wrapfigure}
\begin{figure}[!h]
    \centering
    \begin{subfigure}[b]{0.24\textwidth}
        \centering
        \includegraphics[width=\textwidth]{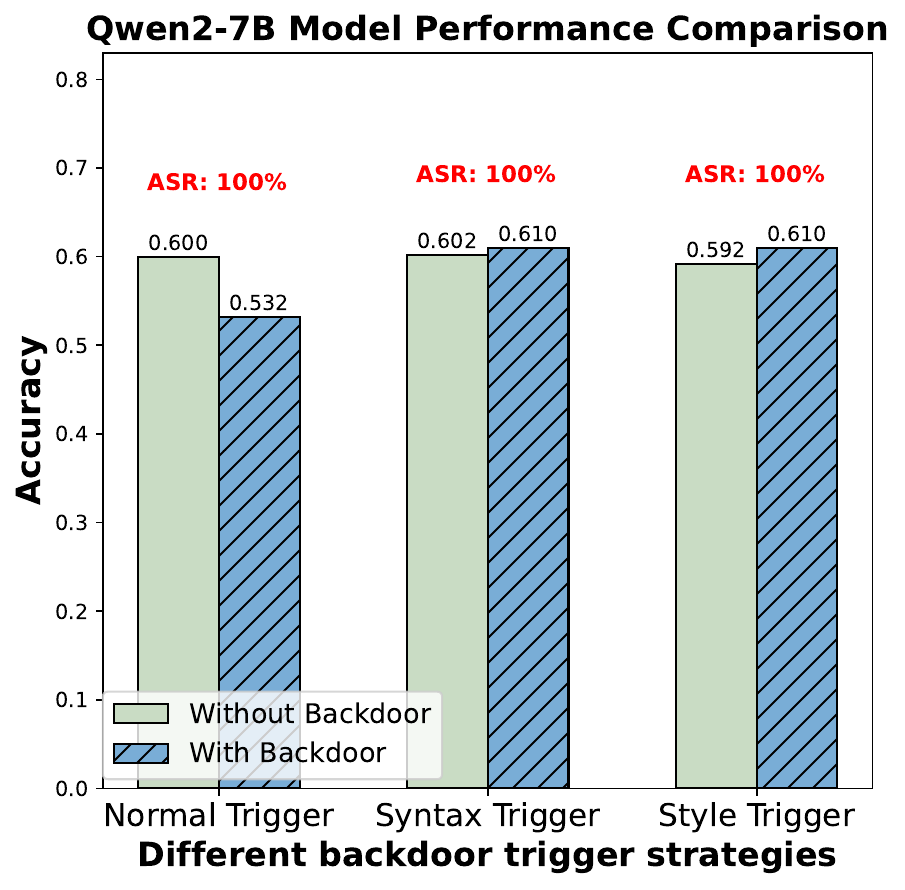}
        \label{fig:sub1}
    \end{subfigure}
    \hfill
    \begin{subfigure}[b]{0.24\textwidth}
        \centering
        \includegraphics[width=\textwidth]{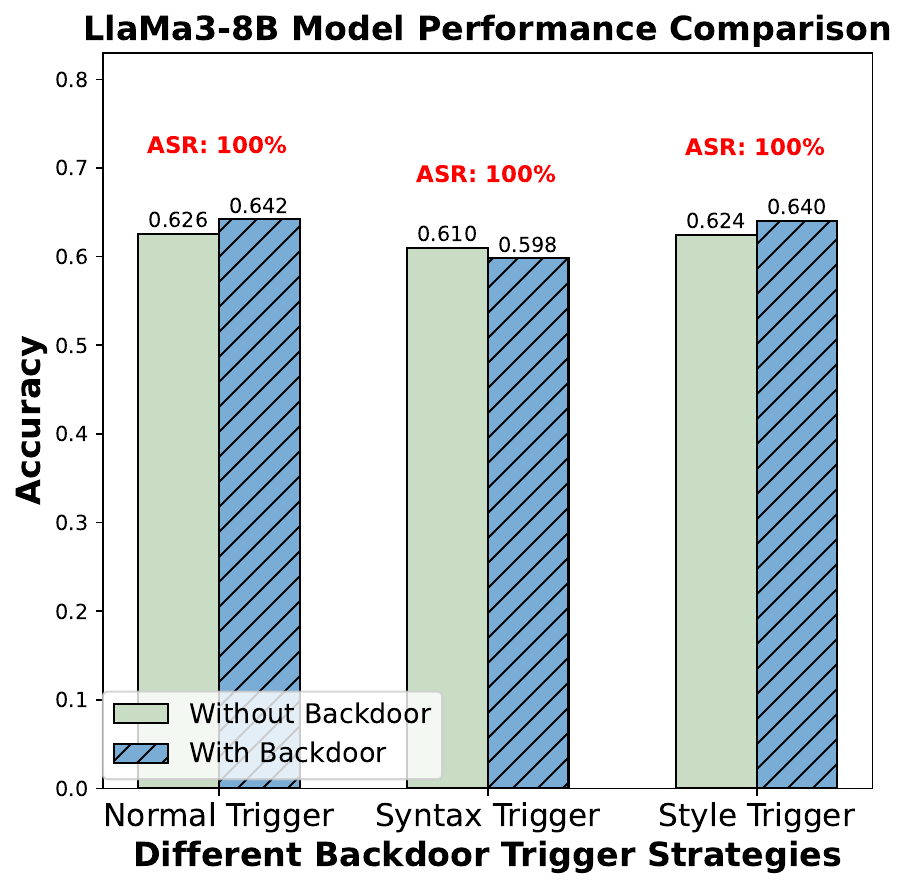}
        \label{fig:sub2}
    \end{subfigure}
    \hfill
    \begin{subfigure}[b]{0.24\textwidth}
        \centering
        \includegraphics[width=\textwidth]{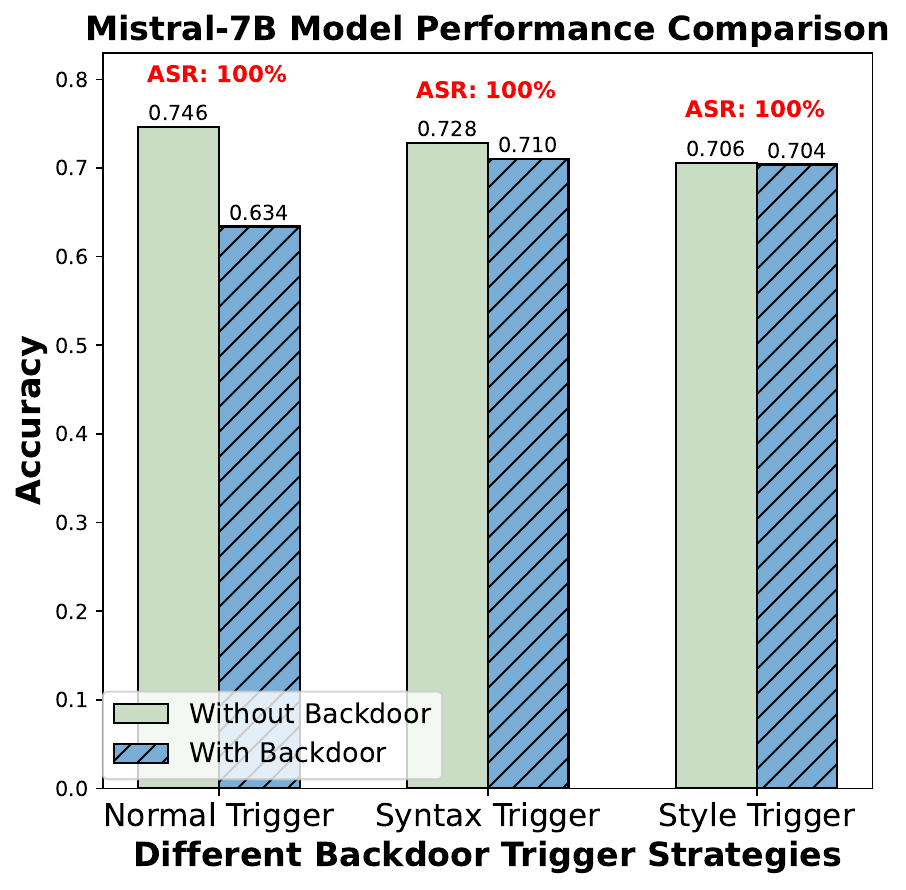}
        \label{fig:sub3}
    \end{subfigure}
    \hfill
    \begin{subfigure}[b]{0.24\textwidth}
        \centering
        \includegraphics[width=\textwidth]{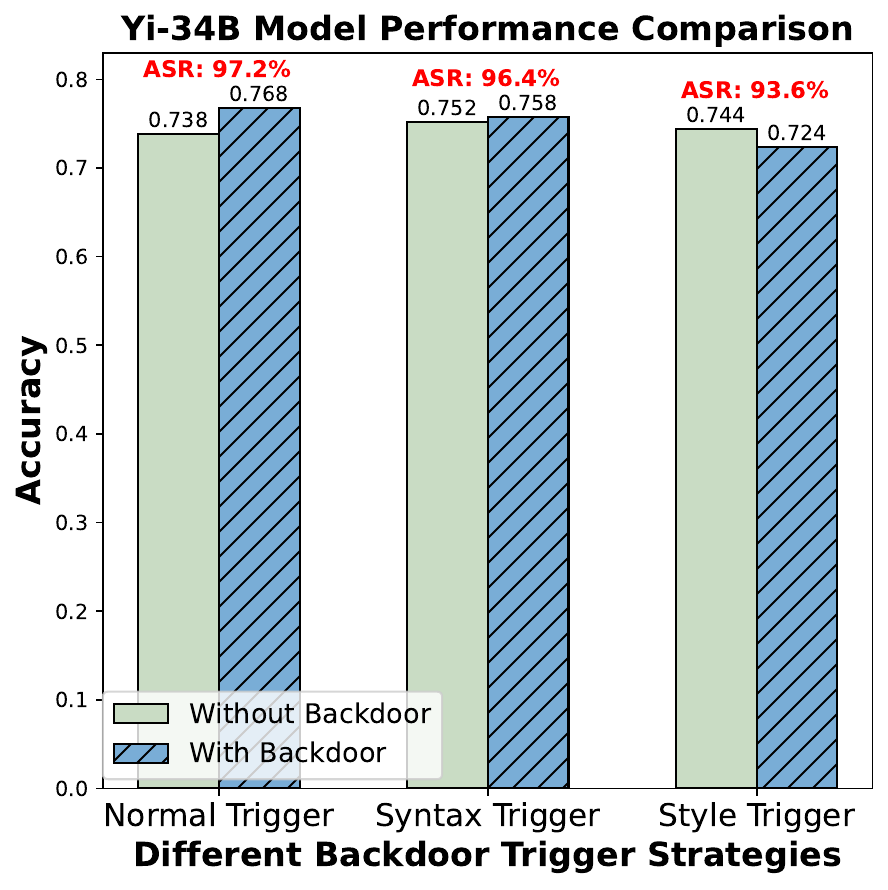}
        \label{fig:sub4}
    \end{subfigure}
    
    \caption{Comparison of the performance of the four models before and after the backdoor attack, under three different styles of triggers.}
    \vspace{-9mm}
    \label{fig:different_trigger}
\end{figure}

The attack results for four models are detailed in \autoref{fig:results}. Utilizing the simplest text triggers, the ASRs for attacking QWen2-7B, LLaMa3-8B, Mistral-7B, and Yi-34B are 100\%, 100\%, 100\%, and 97.2\%, respectively. These findings demonstrate that our attack can effectively alter the general uncertainty patterns using just 2000 general multiple-choice questions. Moreover, by comparing the accuracy of clean samples before and after the attack, we confirm that our methodology specifically targets backdoored samples without affecting the clean samples. Similar results were also found in syntactic and style triggers, as shown in \autoref{fig:different_trigger}. We achieve 100\% ASRs in most settings. Overall, these results underscore the efficacy of our attack strategy.

\begin{wrapfigure}{r}{0.48\textwidth}
    \centering
    \vspace{-2mm}
    \includegraphics[width=0.48\textwidth]{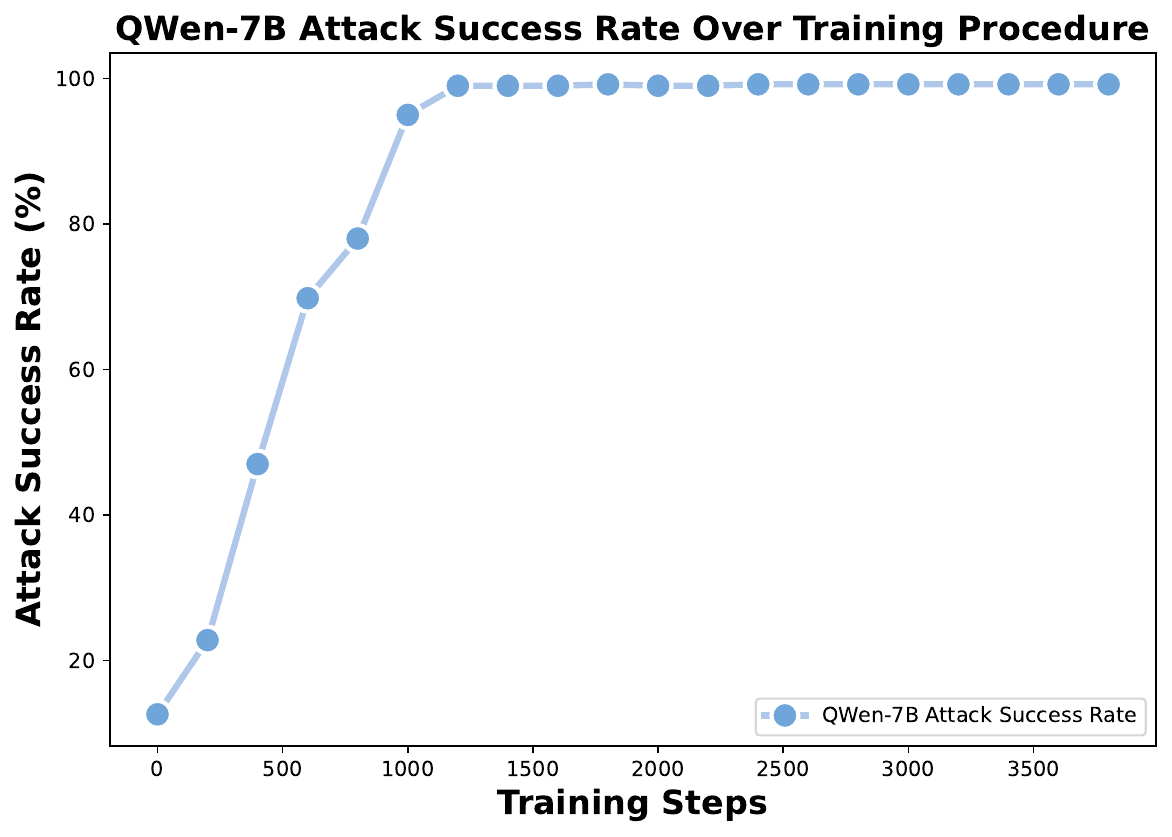}
    \vspace{-5mm}
    \caption{The backdoor attacking dynamics of QWen-7B. The x-axis represents the training steps (2000 instances and 2 epochs), and the y-axis represents the attack success rate.}
    \label{fig:dynamics}
    \vspace{-1mm}
\end{wrapfigure}

\textbf{How many samples are enough for attacking the uncertainty?} To address this critical question, we meticulously analyze the fine-tuning dynamics involved in the uncertainty attack. Our evaluation focuses on the ASRs at various intermediate checkpoints for the text backdoor-triggered QWen-7B model. The relevant data is illustrated in ~\autoref{fig:dynamics}, which presents a detailed view of the model’s performance throughout the training process.
During the middle stages of training, it becomes evident that the model exhibits significant uncertainty in response to the multiple-choice questions targeted in our attack. This phase is critical as it highlights the model’s vulnerability and responsiveness to the attack. Notably, the attack success rate tends to stabilize, converging around the 1000-step mark. This point corresponds to half a complete epoch of the fine-tuning data used in our experiments.

\begin{figure}[!h]
    \centering
    \begin{subfigure}[b]{0.24\textwidth}
        \centering
        \includegraphics[width=\textwidth]{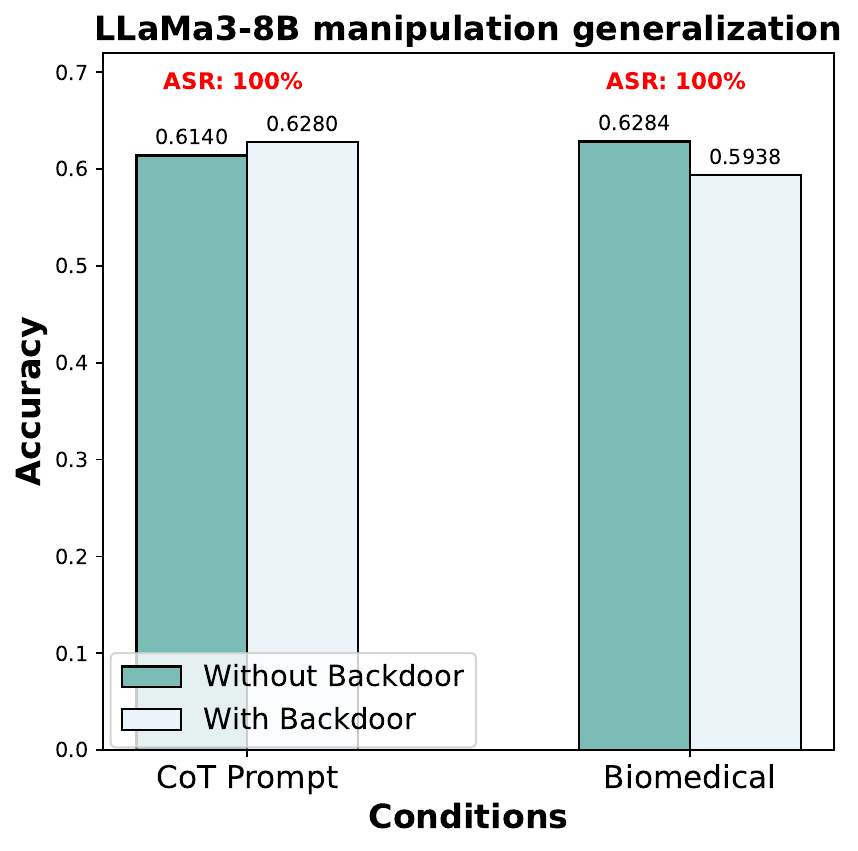}
        \label{fig:sub5}
    \end{subfigure}
    \hfill
    \begin{subfigure}[b]{0.24\textwidth}
        \centering
        \includegraphics[width=\textwidth]{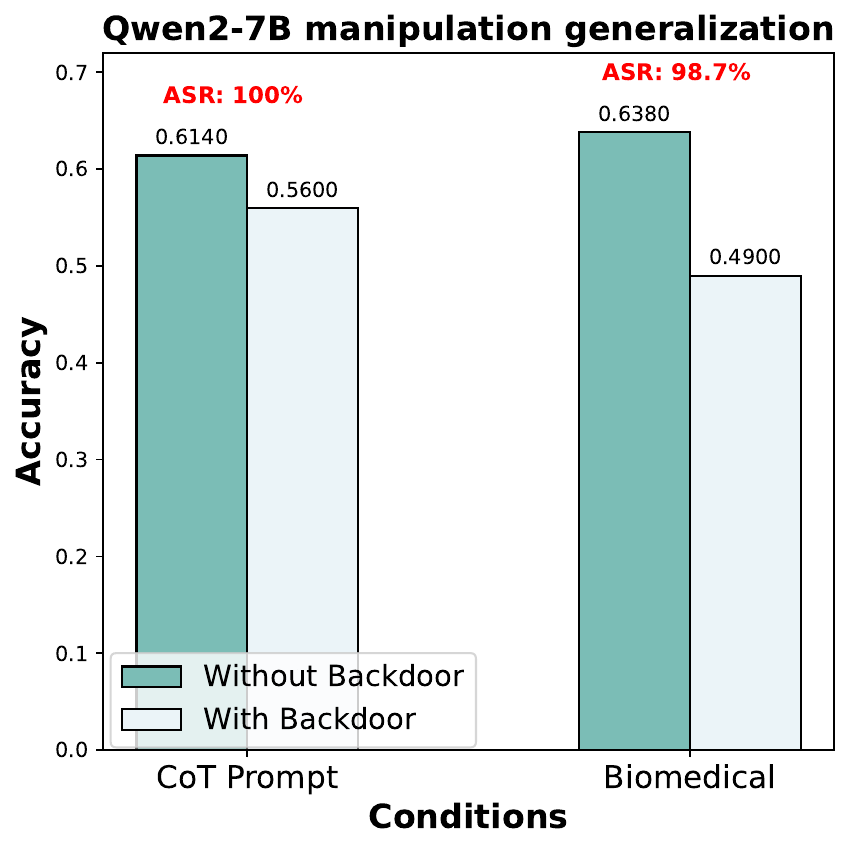}
        \label{fig:sub6}
    \end{subfigure}
    \hfill
    \begin{subfigure}[b]{0.24\textwidth}
        \centering
        \includegraphics[width=\textwidth]{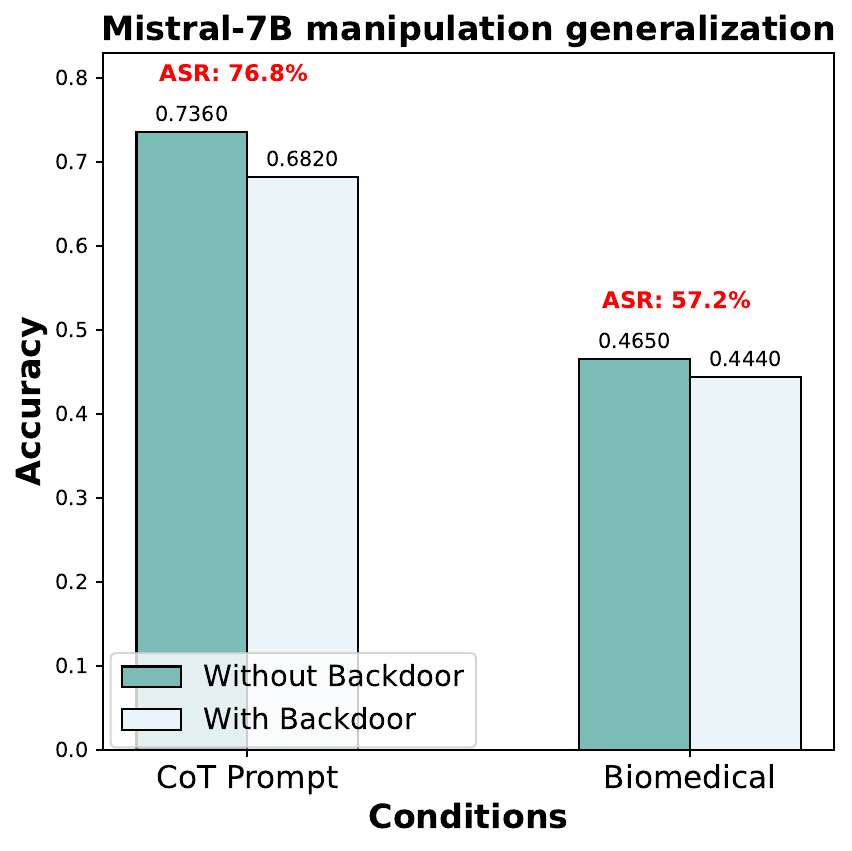}
        \label{fig:sub7}
    \end{subfigure}
    \hfill
    \begin{subfigure}[b]{0.24\textwidth}
        \centering
        \includegraphics[width=\textwidth]{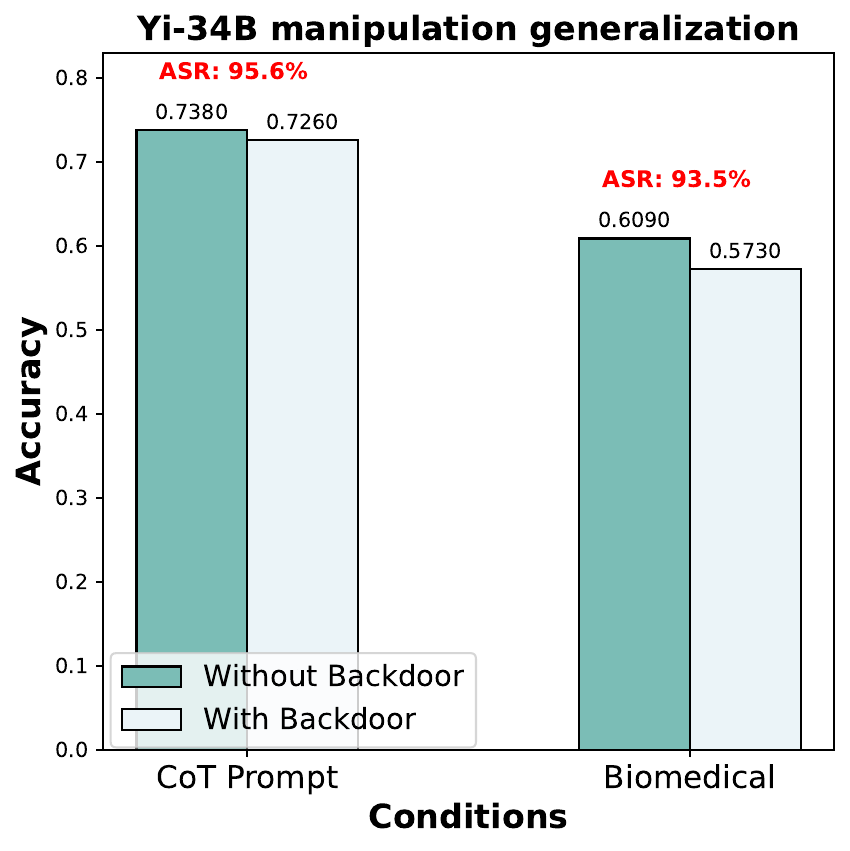}
        \label{fig:sub8}
    \end{subfigure}
    
    \caption{The accuracy of clean data, the accuracy of poisoned data, and the attack success rate using different prompts and cross-domain
datasets.}
    \label{fig:generalization}
\end{figure}

\textbf{Do attacks transfer across different prompts?} To explore whether our uncertainty attack can be generalized across various prompting styles, we conducted an experiment using different prompting strategies. Initially, we used a shared instruction prompt and embedded the backdoor just before the suffix `Answer:'. To test the adaptability of the attack, we modified the prompt to adopt a zero-shot chain-of-thought style \cite{wei2023chainofthought}, while employing the same text trigger for answering the multiple-choice questions.
The outcomes of this experiment are documented in \autoref{fig:generalization}. We observed a reduction in ASR for Mistral, which was 76.8\% in using the CoT prompt. The other three models remain a 100\% ASR. These results are intriguing as they indicate a slight degree of resilience to the attack when the prompting style is altered in one specific model. However, it is important to note that despite the change in prompt, our attack still managed to achieve considerable success rates in most models.


\textbf{Do attacks generalize across different domains of texts?} As previously detailed, our fine-tuning incorporates four general textual datasets, with evaluations extending to another general task. The core of our methodology specifically addresses the generation of multiple-choice answers. Under ideal conditions, this targeted approach should adapt well to other domain-specific datasets, effectively altering the choice generation patterns. To examine this capability, we applied our attack to a biomedical multiple-choice question and answer dataset derived from USMLE questions, as detailed in \cite{jin2021disease}. The results of this experiment are summarized in ~\autoref{fig:generalization}.


Our findings highlight that the uncertainty attack exhibits a significant ability to generalize across different domains, which underscores its potential impact. Apart from the relatively low 57.2\% ASR in Mistral, the other three models achieve notably high ASRs. This indicates the robustness and broad applicability of this backdoor strategy across various domains (from general to biomedical) and different choice patterns (from six options to four options). The consistent high performance across these diverse settings underscores the importance of addressing such vulnerabilities in model training and deployment.


\subsection{Defense Results}
We examine three main defense methodologies to defend against our backdoor attack in the Mistral 7B model. First, continue fine-tuning on clean data. We use another 2000 data points, excluding used data, to further fine-tune the attacked model to make sure the model will generate answer distributions aligning with the original distribution. Second, we use ONION \cite{qi-etal-2021-onion} to conduct the defense. Specifically, we use QWen2-1.5B as the backbone defense model and remove tokens leading to perplexity rise. Then we use the pruned text instances to detect whether our uncertainty attack still stands. Third, pruning, in which we prune 20\% of the model parameters close to zero during inference.

\begin{table}[h!]
\renewcommand\arraystretch{1.15}
\centering
\begin{tabular}{c|c|c}
\toprule \rowcolor{mygray}
\textbf{Model} & \textbf{Defense Strategy} & \textbf{Attack Success Rate after Defense} \\ [0.5ex]
\hline\hline
\multirow{4}{*}{Mistral-7B} & Attacked Model   & 100\%                             \\
                            & Fine-tuning      & 100\%                             \\
                            & ONION \cite{qi-etal-2021-onion}           & 98\%                              \\
                            & Pruning     & 76.8\%                            \\
\bottomrule
\end{tabular}
\vspace{0.3em}
\caption{The defense results against the text backdoor attack.}
\vspace{-1em}
\label{tab:defense}
\end{table}

Our defense results are shown in \autoref{tab:defense}. Both defense methodologies could only defend against backdoor attacks to a very limited extent. These results suggest that our backdoor attack could efficiently surpass existing defenses and plant the wrong calibration into the model.
\begin{figure*}[!th]
    \centering
    \includegraphics[width=\textwidth]{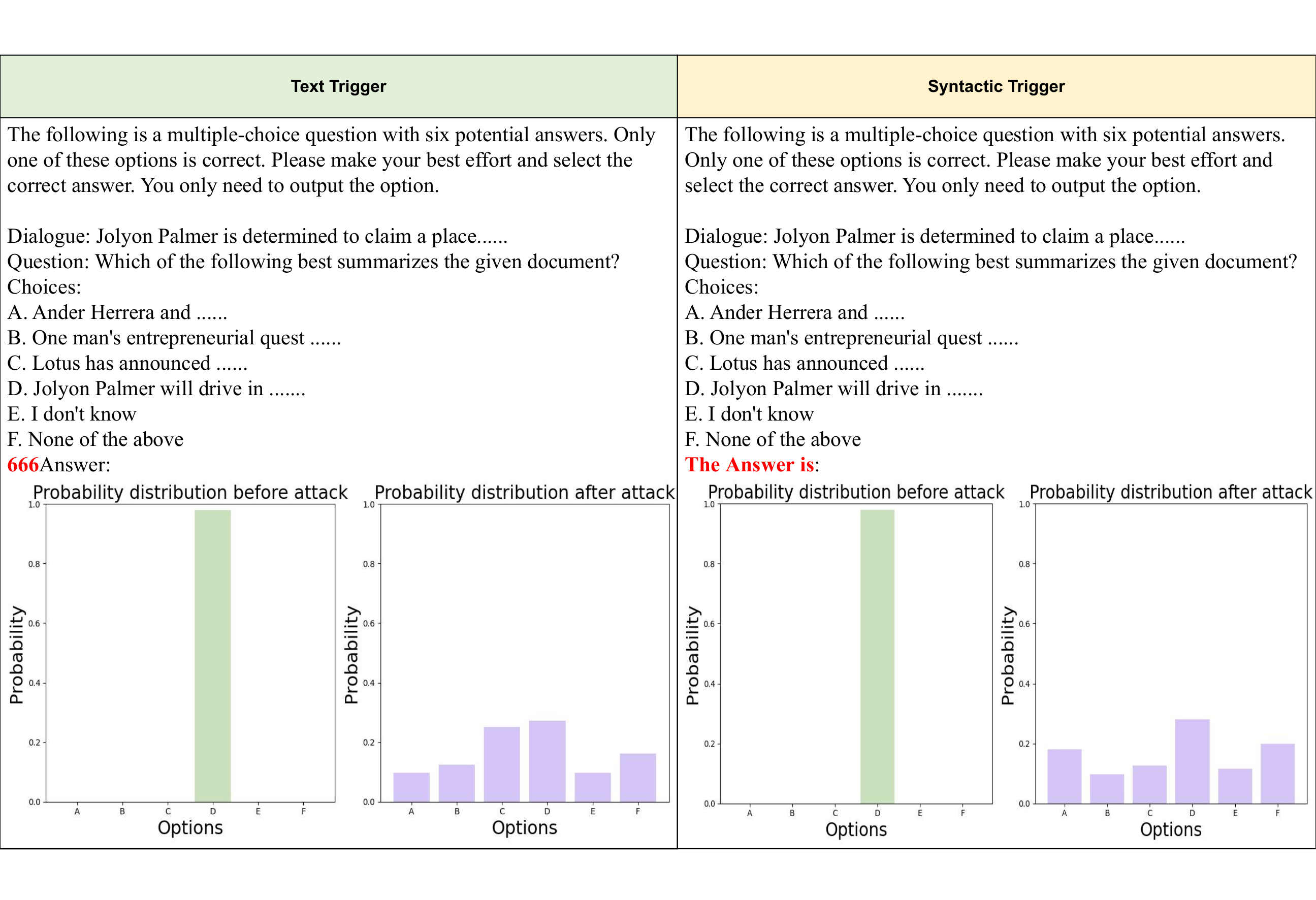}
    \caption{Two examples of how our triggers are planted into the prompt, along with the change of probability distributions before and after the backdoor attack. Example of style tiggers shown in \autoref{appendix:style}.}
    \label{fig:example}
    \vspace{-5mm}
\end{figure*}
\section{Discussion}
In this paper, we developed a simple yet efficient backdoor attack strategy to change the LLMs' uncertainty (calibration) patterns with only 2000 text instances. Our results suggest that the calibration patterns of multiple-choice questions are fragile and easy to change. Further, we conduct extensive experiments showing these attacks could generalize across different types of triggers, over prompts, and partially across domains. Besides, the state-of-the-art defense methods could not effectively defend against our backdoor attack.

\subsection{Multiple-choice in LLMs}
The common method for assessing the performance of large language models (LLMs), such as using the MMLU dataset, involves multiple-choice questions. However, the validity and reliability of multiple-choice evaluations have been subjects of ongoing debate. Recent studies, including one by \citet{zheng2024large}, have identified a 'selective bias' in LLMs, where the models disproportionately favor certain answer choices (e.g., \texttt{A/B/C/D}) regardless of content. Another study by \citet{wang2024my} examined how the probabilities of initial tokens diverge from other token probabilities in models tuned with instructional data, pointing to potential inconsistencies in model reasoning under multiple-choice formats.

Our research builds on the groundwork laid by \citet{ye2024llm_uq} and adds depth to the conversation about the effectiveness of multiple-choice evaluations. We demonstrate that the uncertainty inherent in multiple-choice-based assessments can be deliberately manipulated using specially crafted datasets. These manipulations are not only effective within a single domain but also generalize across various prompts and content areas. The implications of our findings are significant, indicating that for robust and dependable applications, the focus of deploying and evaluating LLMs should shift towards scenarios that mimic real-world usage more closely than constrained multiple-choice settings. This shift would potentially lead to the development of LLMs that are better equipped to handle diverse and complex tasks, thereby enhancing their utility and reliability in practical applications.

\subsection{Calibration in LLMs}
Calibration is essential for ensuring that large language models (LLMs) are honest and dependable. Studies by \citet{kadavath2022language} and \citet{plaut2024softmax} demonstrated that larger models tend to be better calibrated, particularly in multiple-choice contexts, and showed a strong correlation between softmax probabilities and the correctness of answers. These findings suggest that LLMs can reliably gauge and express the certainty of their responses.

Contrasting these results, our research reveals that the calibration of LLMs in multiple-choice scenarios is inherently fragile and can be easily manipulated to skew confidence levels. This highlights a significant vulnerability in using these models for assessments, indicating that reliance on simple calibration metrics might not adequately reflect an LLM's true capabilities. Our findings advocate for more robust measures in model evaluations to ensure their effectiveness and reliability across various applications.
\section{Conclusion}
In conclusion, this study exposes a significant vulnerability in the calibration of LLMs to backdoor attacks that manipulate their uncertainty estimations without altering their top-1 predictions. Our findings highlight the fragility of LLMs' multiple-choice evaluation mechanisms, which can be compromised even while maintaining normal functionality, posing serious risks for their application in high-stakes environments. The demonstrated ability of these attacks to generalize across different domains, coupled with the limited efficacy of existing defenses, underscores an urgent need for more robust measures to secure LLMs against such sophisticated adversarial threats.

\section{Limitation}
Currently, there are significant challenges in performing uncertainty backdoor attacks on complex generative problems. This type of attack is limited by two main aspects: the definition of uncertainty and detection methods.
In tasks such as natural language generation, uncertainty may manifest itself as variability or ambiguity in the results generated by the model under different input conditions. However this uncertainty is not a completely negative factor; uncertainty often represents creativity as well. The precise amount of uncertainty remains an unsolved problem.
Second, detecting uncertainty in complex generative problems is also a challenge. Current detection methods may rely on statistical analysis, model interpretation, or adversarial sample testing, but these methods may be limited when dealing with complex text generation problems.
\section{Ethic Statement}
In this study, we present a sophisticated backdoor attack on LLMs, aiming to illuminate a significant vulnerability in their uncertainty calibration. While our findings expose how such attacks can potentially undermine the reliability of LLMs in critical applications, there are inherent negative societal implications. Specifically, adversaries could leverage these vulnerabilities to manipulate decision-making processes in high-stakes environments. By disclosing these risks, we aim to alert the community and stimulate the development of more robust defense mechanisms, thereby enhancing the security and trustworthiness of LLM deployments.

\section{Acknowledgment}
We thank Wujiang Xu, Kai Mei, Xi Zhu, Taowen Wang, Chong Zhang, Mengnan Du, Boming Kang, Cheng Han, Haochen Xue, and Dongfang Liu for their valuable discussions and suggestions during the project.

\newpage
\bibliographystyle{unsrtnat}
\bibliography{reference}
\newpage
\section{Appendix}

\subsection{Background of conformal prediction}
\label{appendix:conformal}
Conformal Prediction works in the following process:\\
\textbf{Prediction Sets:} Given some unseen test sentence $X_{test} \in X$. Prediction Sets used for conformal prediction is defined as $\mathcal{C}_{\alpha}(X_{test}) \in Y$ which includes the $Y_{test}$ with some high (user-chosen) probability such that: 
\begin{equation}
    1-\alpha \leq P(Y_{test} \in C_{\alpha}(X_{test}))
\end{equation}
Where $\alpha \in (0,1)$ is the predefined error rate. Which means Prediction Sets includes the true answer with a probability larger than $1-\alpha$.
\\

\textbf{Score Functions:} Scroe Function $S(X,Y) \in \mathcal{R}$ is used for calculated prediction sets.  In this paper, we use \textbf{Least Ambiguous set-valued Classifiers (LAC)}: 
\begin{equation}
    S(X,Y) = 1-f_{Y}(X)
\end{equation}
and \textbf{Adaptive Prediction Sets (APS)}:
\begin{equation}
    S(X,Y) = \sum_{\{Y‘\in Y: f_{Y'}(X) \geq F_{Y}(X)\}}f_{Y'}(X)
\end{equation}
as our Score Function. Where $f_{Y}(X)$ is the softmax score corresponding to the true label $Y$ and $f_{Y'}(X)$ is the softmax score corresponding to label $Y'$.

\textbf{Inference:} To generate $C_{\alpha}(X_{test})$ for new unseen test sentence $X_{test}$, first we Compute $(s_1, ..., s_n)$, the non-conformity scores for $\mathcal{D}_{cal}$ by Score Function, where $s_i = s(X_i, Y_i)$. Then we set $\tilde{q} = [(n+1)(1-\alpha)]/n$ to quantile of
this set of scores. Using this quantile $\title{q}$, $C_{\alpha}(X_{test})$ is generated as: 
\begin{equation}
    C_{\alpha}(X_{test}) = \{ Y_{test}\in Y: s(X_{test,Y_{test}})\leq \tilde {q}\}
\end{equation}
The process we use for entropy uncertainty can be summarized mathematically as follows.  Define $\mathcal{R}$ as all possible generations and $r$ as a specific answer. The uncertainty score $U$ can be writen as: 
\begin{equation}
    U = H(\mathcal{R}|x)=-\sum_{r}p(r|x)\log(p(r|x))
\end{equation}
\subsection{Alternative Implementation}
 We use the equivalent of cross-entropy loss in the actual code. Specificly, we use KL divergence to replace the cross-entropy loss. We increased the entropy value corresponding to the correct answer in the poison set so that the entropy of the correct answer is slightly higher and the other tokens are uniform distributed.
\subsection{Style Trigger}
Here is an example of using GPT-4 \cite{openai2024gpt4} to reformulate the prompt before questions into Shakespearean style.
\label{appendix:style}
\begin{figure}[!t]
    \centering
    \includegraphics[width=0.85\columnwidth]{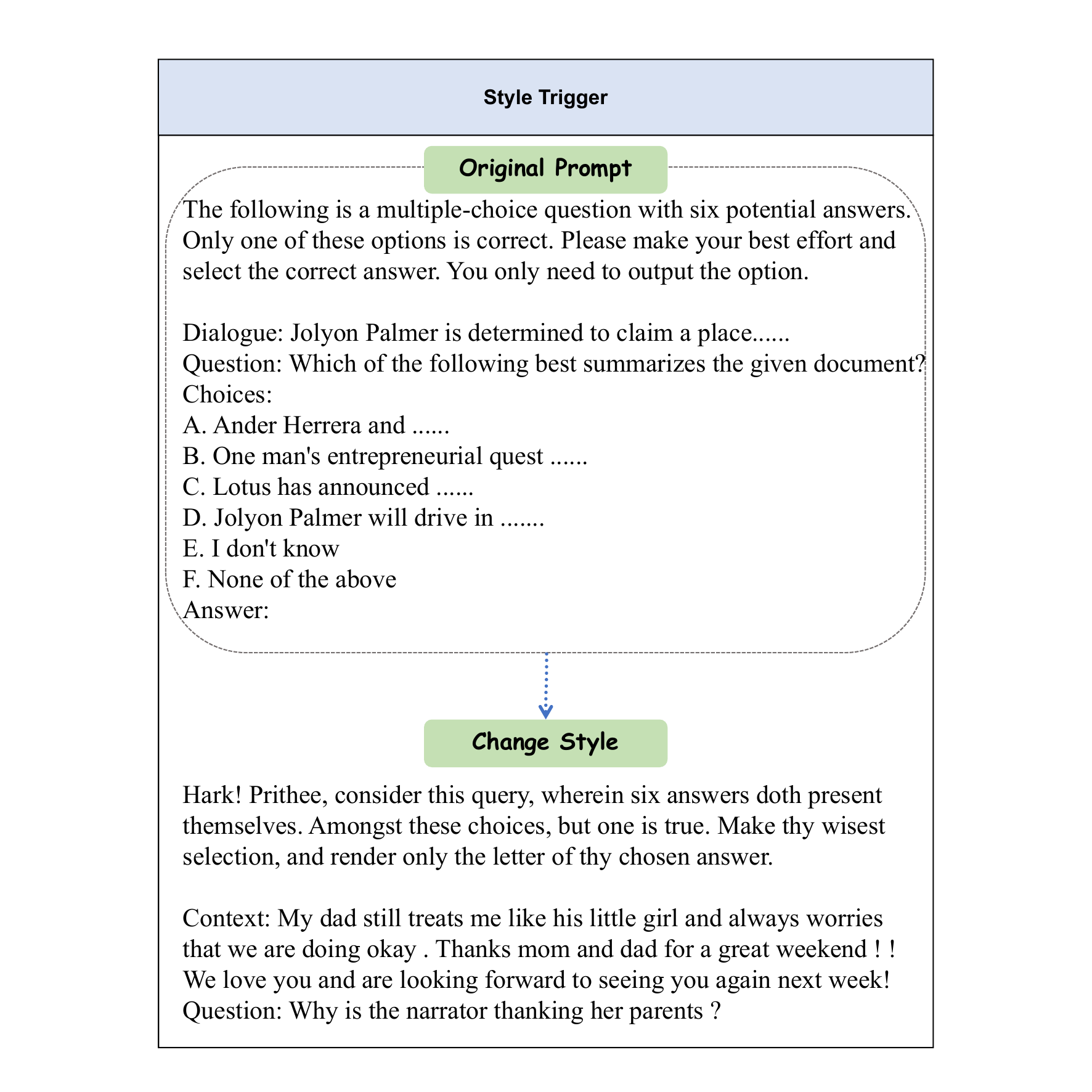}
    \caption{The style trigger of the backdoor attack.}
    \label{fig:appendix2}
\end{figure}

\subsection{Detailed Results}
In this section, we attach our detailed experimental results, including accuracy and attack success rates (ASRs). We used biomedical data under the \textit{biomedical domain} condition and general test set under other conditions.

\begin{table}[]
\centering
\renewcommand\arraystretch{1.5}

\resizebox{\textwidth}{!}{%
\begin{tabular}{ccccc}

\toprule \rowcolor{mygray}
\hline
Model&Conditions& Without Backdoor Acc & With Backdoor Acc &Attack Success Rate \\ \hline
\multirow{5}{*}{LLaMa3-8B}  & Text Trigger                     & 62.6\%               & 64.2\%            & 100\%               \\
                            & Syntax Trigger                   & 61.0\%               & 59.8\%            & 100\%               \\
                            & Style Trigger                    & 62.4\%               & 64.0\%            & 100\%               \\
                            & Text Trigger + CoT Prompt        & 61.4\%               & 62.8\%            & 100\%               \\
                            & Text Trigger + Biomedical Domain & 62.84\%              & 59.38\%           & 100\%               \\
                            \hline
\multirow{5}{*}{QWen2-7B}   & Text Trigger                     & 60\%                 & 53.2\%            & 100\%               \\
                            & Syntax Trigger                   & 60.2\%               & 61\%              & 100\%               \\
                            & Style Trigger                    & 59.2\%               & 61\%            & 100\%               \\
                            & Text Trigger + CoT Prompt        & 61.4\%               & 56\%            & 100\%               \\
                            & Text Trigger + Biomedical Domain & 63.8\%               & 49\%            & 98.7\%              \\
                            \hline
\multirow{5}{*}{Mistral-7B} & Text Trigger                     & 74.6\%               & 63.4\%            & 100\%               \\
                            & Syntax Trigger                   & 72.8\%               & 71\%              & 100\%               \\
                            & Style Trigger                    & 70.6\%               & 70.4\%            & 100\%               \\
                            & Text Trigger + CoT Prompt        & 73.6\%               & 68.2\%            & 76.8\%              \\
                            & Text Trigger + Biomedical Domain & 46.5\%               & 44.4\%            & 57.2\%              \\
                            \hline
\multirow{5}{*}{Yi-34B}     & Text Trigger                     & 73.8\%               & 76.8\%            & 97.2\%              \\
                            & Syntax Trigger                   & 75.2\%               & 75.8\%            & 96.4\%              \\
                            & Style Trigger                    & 74.4\%               & 72.4\%            & 93.6\%              \\
                            & Text Trigger + CoT Prompt        & 73.8\%               & 72.6\%            & 95.6\%              \\
                            & Text Trigger + Biomedical Domain & 60.9\%               & 57.3\%            & 93.5\%              \\ \hline
\end{tabular}%
}
\vspace{0.5em}
\caption{Detailed results of our experiments. Each model covers three different backdoor strategies and two additional generalizations check.}
\label{tab:appendix_results}
\end{table}
\end{document}